# A Tribe Competition-Based Genetic Algorithm for Feature Selection in Pattern Classification

Benteng Ma [1], Yong Xia [1,2*]

1. Shaanxi Key Lab of Speech & Image Information Processing (SAIIP), School of Computer Science, Northwestern Polytechnical University, Xi'an710072, China
2. Centre for Multidisciplinary Convergence Computing (CMCC), School of Computer Science, Northwestern Polytechnical University, Xi'an710072, China
Email: yxia@nwpu.edu.cn

Abstract—Feature selection has always been a critical step in pattern recognition, in which evolutionary algorithms, such as the genetic algorithm (GA), are most commonly used. However, the individual encoding scheme used in various GAs would either pose a bias on the solution or require a pre-specified number of features, and hence may lead to less accurate results. In this paper, a tribe competition-based genetic algorithm (TCbGA) is proposed for feature selection in pattern classification. The population of individuals is divided into multiple tribes, and the initialization and evolutionary operations are modified to ensure that the number of selected features in each tribe follows a Gaussian distribution. Thus each tribe focuses on exploring a specific part of the solution space. Meanwhile, tribe competition is introduced to the evolution process, which allows the winning tribes, which produce better individuals, to enlarge their sizes, i.e. having more individuals to search their parts of the solution space. This algorithm, therefore, avoids the bias on solutions and requirement of a pre-specified number of features. We have evaluated our algorithm against several state-of-the-art feature selection approaches on 20 benchmark datasets. Our results suggest that the proposed TCbGA algorithm can identify the optimal feature subset more effectively and produce more accurate pattern classification.

## 1. Introduction

Numerical features, a bridge between input and models, play a pivotal role in data mining and pattern recognition. Although more features are expected to have more discriminatory power, a large number of features may contain a lot of redundancy, and hence significantly degrade the accuracy and generalization of learned models as well as the learning speed, which is widely known as the curse of dimensionality [1]. Therefore, an indispensable step in data mining and pattern recognition procedures is dimensionality reduction, which can be roughly categorized into feature transformation and feature selection.

Feature transformation targets at projecting features from a high-dimensional space into a low-dimensional space [2]. Well-known feature transformation algorithms include the principal component analysis (PCA) [3], independent component analysis (ICA) [4], linear discriminant analysis (LDA) [5] and their variants. Although being able to reduce the dimensionality of features, these algorithms may destroy the physical meaning of each feature component during the transformation, which complicates further analysis of the model and makes it less interpretable [6]. Alternatively, feature selection aims to choose a subset of available features that are associated with the response variable by excluding relevant and redundant features [7]. Since it can reduce the dimensionality of features while keeping the physical meaning of each feature component, feature selection has distinct advantages over feature transformation in terms of model readability and interpretability.

During the past decades, many feature selection algorithms have been proposed in the literature, which can be divided into three categories [1]: filter methods[8], embedded methods [9] and wrapper methods [10]. In filter methods, selecting or removing a feature component is decided by a criteria function, such as the mutual information, interclass distance or statistical measures. In spite of their computational efficiency, filter methods usually have limited accuracy, due to the absence of optimizing the performance of any specific classifiers directly. On contrast, wrapper methods use the classification performance of a specific classifier to assess the discriminatory power of candidate feature subsets, and thus identify the optimal feature subset with respect to the classifier. However, since the classifier has to be trained by using each selected subset of features, respectively, wrapper methods have intrinsically higher complexity. As the special cases of wrapper methods, embedded methods are characterized by a deeper interaction between feature selection and classifier construction. In these methods, the optimal feature subset is generated while the classifier is constructed.

Since the performances of features and classifiers depend mutually on each other, wrapper methods have been widely used, in which the strategy for searching the optimal feature subset can be either greedy or stochastic. Two of the most classical wrapper methods with the greedy search strategy are the sequential forward selection (SFS) [11] and sequential backward selection (SBS) [12], which, however, suffer from the nesting effect, i.e. the feature that is selected or removed cannot be removed or selected in subsequence steps. This issue can be alleviated by jointly using SFS and SBS. A typical example is the "plus $l$ take away $r$" method [13], which enlarges the feature subset by adding $l$ features using SBS and then deletes $r$ features using SBS. To avoid the difficulty of determining appropriate values for $l$ and $r$, Pudil et al. [14] proposed the sequential backward floating selection (SBFS) and sequential forward floating selection (SFFS) algorithm, in which the values of $l$ and $r$ are updated adaptively.

Since greedy search makes local decisions, stochastic search should be employed to identify the globally optimal feature subset. Shehata and White [15] proposed a randomization method to assess the statistical significance for best subset regression. Despite the method corrected a non-trivial problem with Type I errors, it would still take into account the number of potential features and the inter-dependence between features. However, Evolutionary algorithms, such as the genetic algorithm (GA), genetic programming (GP), ant colony optimization (ACO) and particle swarm optimization (PSO), have proven performance in finding optimal solutions for complex and nonlinear problems [6] with neither prior domain knowledge nor a differentiable objective function, and hence are very suitable for solving the feature selection problem. Muni et al. [16] assumed that each classifier has $c$ trees, where $c$ is the number of classes, and developed a multi-tree GP algorithm for simultaneous feature selection and training a classifier. Sheikhpour et al. [10] proposed the PSO-KDE model, which combines PSO with a non-parametric kernel density estimation (KDE) based classifier to distinguish benign breast tumors from malignant ones. In this model, PSO simultaneously optimizes the selected feature subset and the kernel bandwidth in the KDE-based classifier. Jensen and Shen [17] applied ACO to searching a feature subset in a rough set and achieved good results on a relatively small set of features. The comparative study conducted by Santana et al. [18] shows that, if the number of features is small, ACO performs better than other evolutionary approaches; otherwise, GA performs better.

**1.1 Related Work**

GA is most likely the first evolutionary computation technique that has been applied to feature selection [19]. Kabir et al. [20] incorporated local search operations into GA and utilized the correlation information in conjunction with the bounded scheme to select a subset of salient features. Li et al. [21] proposed a bi-encoding scheme-based GA to select features for image annotation, in which each individual consists of a pair of strings, a binary string encoding the selection of features and a real valued string indicating the weights of selected features. Hamdani et al. [22] proposed a bi-coded representation in GA for feature selection. They encoded each individual with two chromosomes: a binary chromosome representing the presence of features in the candidate solution and a real-valued chromosome representing the confidence rates of features, which are used to assign different weights to features during the classification procedure. Chen et al. [23] developed a GA-based approach to feature selection and classification of microarray data, in which each individual has two parts: an integer that represents the number of selected features and an integer string that gives the selected feature components. Yahya et al. [24] explored variable length representation of individuals, in which each individual gives the selected features only and different individuals may have different lengths, and thus developed an evolutionary filter approach to feature selection.

It is commonly acknowledged that the encoding scheme for individuals plays an important role in GA-based feature selection. Above-mentioned approaches mainly employ two types of encoding schemes: binary encoding and integer encoding [23]. Supposing selecting an optimal feature subset from $N$ features, binary encoding defines each individual as a $N$-bit binary string, where "1" shows the corresponding feature is selected and "0" means discarded, and integer encoding defines each individual as an integer string with the length of selected features, where each integer gives a selected feature component. Binary encoding does not pose any constraints to the number of selected features. When selecting $m$ feature components, we have $C_N^m$ possible solutions. It is obvious that $C_N^{m_1} \gg C_N^{m_2}$, when $m_1 \cong \frac{N}{2}$ and $m_2$ is a very small or very large number. That means the vast majority of possible solutions in the search space contains about $\frac{N}{2}$ selected feature components. Thus the binary encoding is prone to resulting in a larger feature subset than the optimal one. The integer encoding can not only avoid this bias, but also has a fixed individual size that does not enlarge with the increase of candidate features. However, this scheme requires a pre-determined number of selected feature components, which is usually hard to estimate.

Besides the encoding scheme, multi-population techniques have been widely investigated in recent GA-based feature selection studies. Although conventional GA has one single population, it has recently been shown that better results can be achieved by introducing multiple parallel populations [25]. Chang et al. [26] proposed a two-phase sub-population GA. In the first phase, the population is decomposed into many sub-populations, which are independent and unrelated to each other, and each sub-population is fixed for a pre-determined criterion. In the second phase, non-dominant solutions are combined and all sub-populations are unified as one big population. As Affenzeller [27] suggested that sub-populations should communicate to each other to bring better convergence and diversity and eventually to further improve the solution, they extended their two-phase method to the SPGA II algorithm [28], which introduces the mechanism to exchange information among sub-populations. Once a sub-population reaches a better non-dominated solution, other sub-populations are able to apply it directly within their searching areas. The idea of this mechanism is to share the Pareto set generated by different sub-populations and to save these Pareto sets as a global archive, which will guide all individuals in the same population searching toward the true Pareto front. However, it is difficult to know whether the decomposition is reasonable and an inappropriate decomposition may result in bad performance. To further improve the interaction and cooperation among sub-populations, Li et al. [29] proposed a multi-population agent GA with a double chain-like agent structure for feature selection, in which every sub-population is connected to each other with one cycle chain and shares two common agents. Due to the shared agents, sub-populations can exchange genetic information with each other to explore the optimal solution. Different from the sharing strategy, Pourvaziri et al. [25] proposed a hybrid multi-population GA, in which multiple sub-populations first evolve independently and then are combined to form a main population that continues to evolve until the stopping criterion is met. In this way, the various parts of the solution space are most likely explored. In multi-population GAs, sub-populations can optimize different objectives [30]. Derrac et al. [31] proposed a cooperative co-evolutionary algorithm for feature selection, in which the GA has three sub-populations, one focusing on feature selection, one on instance selection and the other one on both feature and instance selection.

**1.2 Outline of Our Work**

In this paper, we propose a tribe competition-based genetic algorithm (TCbGA) for feature selection in pattern classification, which takes the advantages of binary and integer coding schemes as well as the multi-population technique. The population of individuals is divided into multiple tribes. Initialization and evolutionary operations are designed to ensure that the number of individuals, which select a specific number of features, follows a Gaussian distribution $\mathcal{N}(\mu_k, \sigma)$ in each tribe $T_k$. Thus the tribe $T_k$ is mainly responsible for exploring a part of the solution space, where the number of selected features ranges from $\mu_k - 3\sigma$ to $\mu_k + 3\sigma$. Since the features that make up the global optimum must exist in the subspaces searched by one or two elite tribes, inter-tribe competition is introduced to the evolution to predict the elite tribe. The size of the predicted elite tribe is enlarged to give it more search power, and the size of worst-performed tribe is reduced to keep the population size unchanged. This penalty and award strategy enables the algorithm not only to search the solution space locally, but also to quickly look for the global optimal. We have evaluated the proposed algorithm against the state-of-the-art feature selection methods on 20 benchmark datasets.

## 2. Algorithm

As a heuristic-guided parallel and stochastic search approach, TCbGA searches a global optimal subset of features from $N$ candidates through evolving a population of $N_P$ individuals. Each individual encodes a feature selection scheme using binary coding and has a fitness that is defined as the classification accuracy obtained by applying those selected features to the linear SVM [32] in 10-fold cross validation.

Since it is hard to estimate how many features are selected in the global optimal solution and, without this value, traditional binary-coded GA is prone to select about $N/2$ features, we divide the solution space into $N_T$ partly overlapped subspaces and, accordingly, divide the population evenly into $N_T$ tribes, each exploring one subspace. Those tribes evolve in a two-step iterative way. In the intra-tribe evolution step, each tribe evolves independently and the highest fitness in it improves gradually as new generations of the tribe are iteratively produced by using the selection, crossover, and mutation operations, which mimic the genetic processes of biological organisms, such as reproduction, natural selection and natural mutation. In the inter-tribe competition step, different tribes compete against each other by comparing the best individual in each of them. As a result, the tribes that produce better individuals win the right to enlarge their size, and thus obtain the privilege to have more individuals to search their part of the solution space; whereas other tribes have to reduce their size to keep the total number of individuals in the entire population unchanged. This reproduction and competition process goes through one generation to another, until it converges when the highest fitness is constant for many generations or the required number of generations $N_G$ is reached. The diagram of the proposed algorithm is summarized in Fig. 1.

### 2.1 Initialization

Let $\Omega_m$ be the assembly of all admissible individuals, which select $m$ feature components. TCbGA aims to identify the individual with highest fitness from the solution space $\Omega = \bigcup_{m=1}^{N} \Omega_m$ by simultaneously evolving $N_T$ tribes.

In the $k$-th tribe $T_k$, the number of individuals belonging to $\Omega_m$, can be calculated as

$$n_{km} = |T_k \cap \Omega_m| \tag{1}$$

where $|\cdot|$ gives the cardinality of a set. We assume that $n_{km}$ follows a Gaussian distribution $\mathcal{N}(\mu_k, \sigma^2)$ and $\{\mu_1, \mu_2, \cdots, \mu_{N_T}\}$ equally distribute in the range $[1, N]$. The mean value $\mu_k$ largely determines which part of the solution space is explored by the tribe $T_k$, and the standard deviation $\sigma$ governs the searching scope of each tribe.

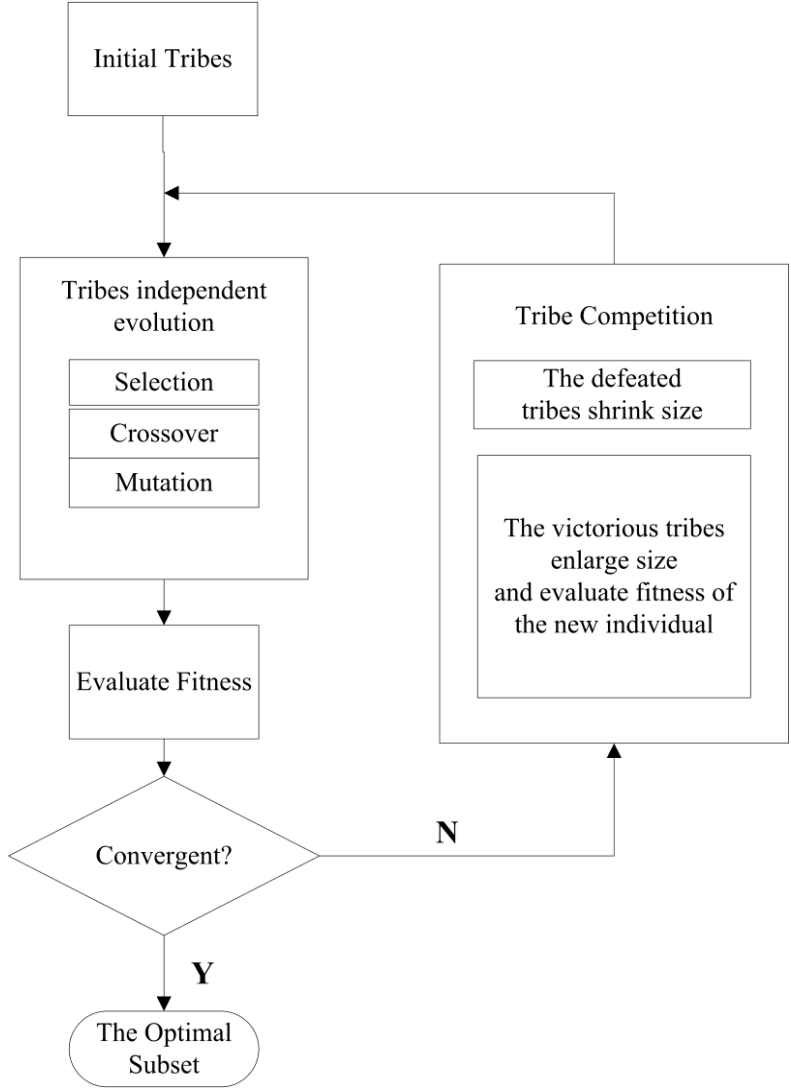

Fig. 1. Diagram of the proposed algorithm

During initialization, we generate random individuals and select some of them form the tribe $T_k$ by ensuring

$$n_{km} = round\left(N_{Tk} \frac{\mathcal{N}(m;\ \mu_k, \sigma^2)}{\sum_{i \in \aleph} \mathcal{N}(i;\ \mu_k, \sigma^2)}\right) \tag{2}$$

where $round(\cdot)$ is the nearest integer function, $\mathcal{N}(m;\ \mu_k, \sigma^2) = \frac{1}{\sigma\sqrt{2\pi}} exp\left[-\frac{(m-\mu_k)^2}{2\sigma^2}\right]$, and $N_{T_k}$ is the number of individuals in $T_k$ and is initially set to $\frac{N_P}{N_T}$.

Obviously, $n_{km}$ takes a large value if $m$ approaches to $\mu_k$, and a small value otherwise. Such bias enables the tribe $T_k$ to be mainly responsible for exploring the subspace $\Omega_{\mu_k}$ and its adjacent subspaces. However, there might not be enough individuals to search the subspace $\Omega_m$, which is far away from $\Omega_{\mu_k}$. To compensate this bias, we let any two adjacent tribes be half-overlapped. Thus each tribe $T_k$ explores $\frac{2}{(N_T+1)}$ of the solution space. A typical example is shown in Fig. 2. Readers are referred to the discussion section for details on the settings of the standard deviation $\sigma$, population size $N_P$ and number of tribes $N_T$.

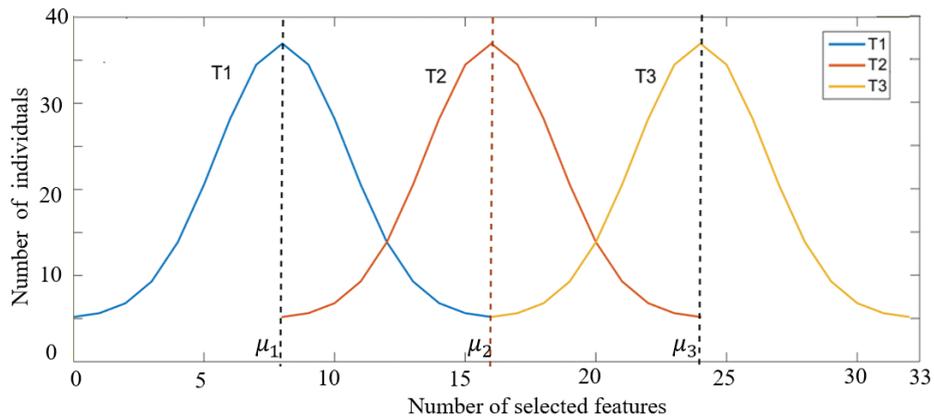

Fig. 2. An example of the distribution of individuals which selected different numbers of features in three tribes (denoted by T1, T2 and T3).

**2.2 Intra-Tribe Evolution**

The intra-tribe evolution, including selection, crossover and mutation, is similar to that of traditional GA, except for keeping $n_{km}$ as fixed as possible to ensure its Gaussian assumption is satisfied.

First, we use an elitism roulette wheel selection scheme based on the rank of fitness to perform selection. The individuals with high fitness have more chance to be selected even tautologically as parent chromosomes for the next generation. $N_{Tk}$ individuals are selected after this step in $T_k$.

Then, segments of the two parent individuals are exchanged during crossover for generating better individuals. Each individual in the original Gaussian tribe would take crossover with the one generated in the selected step. To ensure that $n_{km}$ is unchanged, the crossover between two parent individuals can be expressed by swapping the same number of non-zero elements in their strings.

Fig. 3(a) shows a typical example, in which two parent individuals $\chi_i = $ "1011001100" and $\chi_j = $ "0100110000" represent the selected feature set $F_i = \{f_1, f_3, f_4, f_7, f_8\}$ and $F_j = \{f_2, f_5, f_6\}$, respectively. The crossover position in $\chi_i$ is located between the third and fourth bits. Consequently, the selected feature set $F_i$ is divided into two subsets $F_{iF} = \{f_1, f_3\}$ and $F_{iR} = \{f_4, f_7, f_8\}$. To ensure the same number of selected features to be swapped, the crossover position in $\chi_j$ must locate between the fifth and sixth bits, resulting in $F_{jF} = \{f_2, f_5\}$ and $F_{jR} = \{f_6\}$. The crossover operator swaps the segments of both individuals. As a result, two newly generated children individuals are $\chi'_i = $ "0101101100" and $\chi'_j = $ "1010010000", representing the selected feature sets $F'_i = F_{jF} \cup F_{iR} = \{f_2, f_5, f_4, f_7, f_8\}$ and $F'_j = F_{iF} \cup F_{jR} = \{f_1, f_3, f_6\}$. Let the number of selected feature components in the individual $\chi_i$ be denoted by $n(\chi_i)$. We have $n(\chi_i) = n(\chi'_i) = 5$ and $n(\chi_j) = n(\chi'_j) = 3$. Therefore, $n_{k5}$ and $n_{k3}$ in tribe $T_k$ keep unchanged after this crossover operation.

In the above example, $F_{iF} \cap F_{jR} = \emptyset$ and $F_{jF} \cap F_{iR} = \emptyset$. However, if any of those two intersections is not empty, a children individual will select less feature components than its corresponding parent, due to the duplication caused by the crossover. Fig. 3(b) shows a typical example of this case, where $F_{jF} \cap F_{iR} = \{7\} \neq \emptyset$. Then, after the crossover $|F'_i| = |F_i| - 1$. To keep the number of selected features unchanged, we force one bit in the child individual $F'_i$ to reverse from 0 to 1. Alternatively, when $F_{iF} \cap F_{jR} \neq \emptyset$, we have $|F'_j| < |F_j|$ and have to force one or more bits in the child individual $F'_j$ to reverse from 0 to 1.

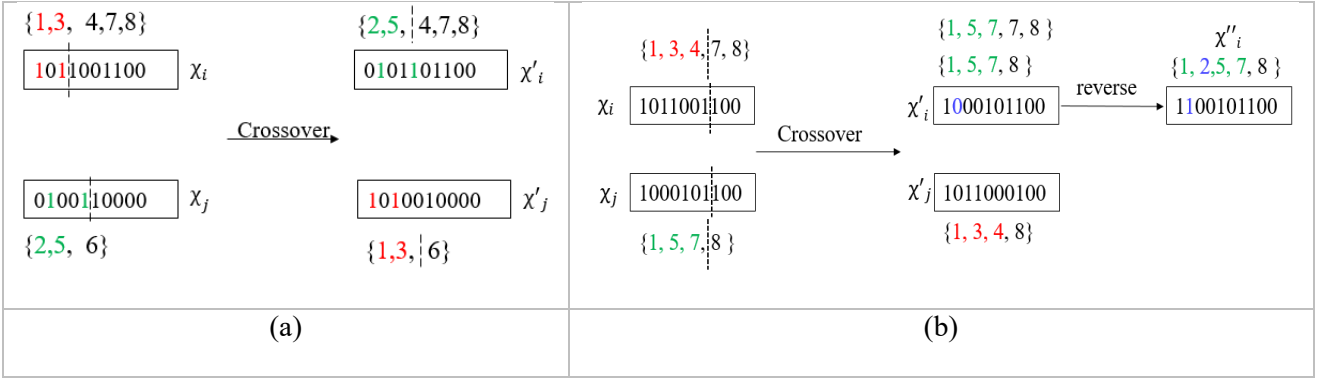

Fig. 3. Example of the crossover process in the proposed algorithm: $\chi_i$ and $\chi_j$ are two parent individuals, and $\chi'_i$ and $\chi'_j$ are two children individuals generated by the crossover.

Next, the mutation operation is applied to individuals to produce sporadic and random alteration of the bits of strings, which can bring the diversity of the species. When a bit of the individual $\chi_i$ in the tribe $T_k$ mutates from 0 to 1, a new feature component is added to the subset specified by $\chi_i$, and hence $n(\chi'_i) = n(\chi_i) + 1$, which leads to

$$\begin{cases} n'_{km} = n_{km} - 1 \\ n'_{k(m+1)} = n_{k(m+1)} + 1 \end{cases} \quad (3)$$

where $m = n(\chi_i)$. To keep the Gaussian distribution of $n_{km}$ unchanged, we randomly choose an individual $\chi_j \in T_k \cap \Omega_{m+1}$ and mutate one of its bits from 1 to 0, shown as in Fig. 4. Similarly, when a bit of the individual $\chi_i$ mutates from 1 to 0, we randomly choose an individual $\chi_j \in T_k \cap \Omega_{m-1}$ and mutate one of its bits from 0 to 1.

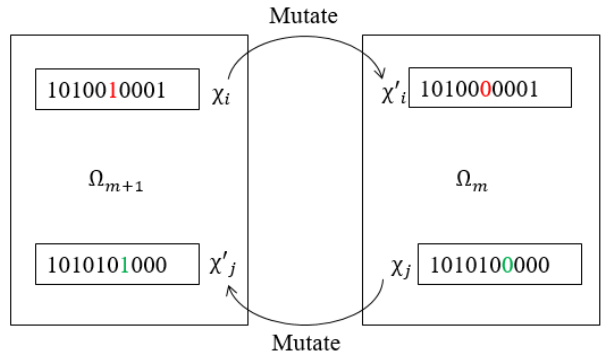

Fig. 4. Example of the mutation process in the proposed algorithm: $\chi_i$ and $\chi_j$ are two individuals before mutation, and $\chi'_i$ and $\chi'_j$ are two mutated individuals.

After selection, crossover and mutation, a new generation of tribe is produced with the same number of individuals and the same Gaussian distribution of $n_{km}$. It should be noted that, to ensure the highest fitness in each tribe increases monotonously during the evolution, we directly inherit the elitist, which has the highest fitness, to the next generation to replace the individual with the lowest fitness in the corresponding sub-solution space.

**2.3 Inter-Tribe Competition**

Tribe competition occurs after all tribes independently evolve two generations, and is revealed by the change of the sizes of tribes. In this step, we gather the elitists of all tribes and sort them by ranking their fitness. The tribes with high-ranking elitists are more likely to be exploring the sub-solution spaces, where the global optimum lies, and hence should be awarded with more searching resources. On the contrary, the tribes with low-ranking elitists are not likely to work in the right places, and hence should release some of their resources to those capable ones. There are different strategies to adjust the size of tribes according to the competition result. In this study, we choose to enlarge the size of the tribe with the best elitist by one and accordingly to shrink the size of the tribe with the worst elitist by one.

Specifically, one individual is added to the enlarged tribe and one individual is discarded in the shrunk tribe, which leads to

$$\begin{cases} N'_{Tk} = N_{Tk} \pm 1 \\ n''_{km} = \left\lfloor \frac{\mathcal{N}(m;\ \mu_k, \sigma^2)}{\sum_{m=1}^{N} \mathcal{N}(m;\ \mu_k, \sigma^2)} * N'_{Tk} \right\rfloor \end{cases}. \tag{4}$$

The update of $n''_{km}$ from the $n_{km}$ is smooth and the individuals would be fine-tuned according to $c_{km}$. $c_{km}$ can be calculated as

$$c_{km} = n''_{km} - n_{km}. \tag{5}$$

When $c_{km}$ is positive, $T_k$ would add $c_{km}$ individuals $\{\chi_1, \chi_2, ..., \chi_{c_{km}}\} \in \Omega_m$ randomly. When $c_{km}$ is negative, $T_k$ would cut down $-c_{km}$ individuals $\{\chi_1, \chi_2, ..., \chi_{-c_{km}}\} \in T_k \cap \Omega_m$.

After the tribe competition, tribes continue to evolve independently until the next tribe competition occurs or the convergence is reached.

## 3. Experiments and Results

The proposed algorithm was evaluated against several state-of-the-art feature selection methods on 20 benchmark datasets acquired from the UCI Machine Learning Repository [33]. These datasets have diversified physical background, number of features, number of classes and number of instances, representing a variety of pattern classification problems. The detailed descriptions of the datasets are shown in Table 1.

Table 1. List of 20 datasets used in this study

|  | Dataset | No. of Classes | No. of Features | No. of Instance |
|---|---|---|---|---|
| Bi-class datasets | WBCD | 2 | 9 | 699 |
|  | Heart | 2 | 13 | 270 |
|  | Australian | 2 | 14 | 690 |
|  | German | 2 | 21 | 1000 |
|  | WDBC | 2 | 30 | 569 |
|  | Ionosphere | 2 | 34 | 351 |
|  | KR vs KP | 2 | 36 | 3196 |
|  | Spam Base | 2 | 57 | 4601 |
|  | Sonar | 2 | 60 | 208 |
| Multiple-class datasets | Wine | 3 | 13 | 178 |
|  | Zoo | 7 | 16 | 101 |
|  | Waveform | 3 | 21 | 5000 |
|  | Lung | 3 | 56 | 32 |
|  | Vehicle | 4 | 18 | 846 |
|  | Dermatology | 6 | 33 | 366 |
|  | Arrhythmia | 16 | 279 | 452 |
| High-dimensional datasets | Hill-Valley | 2 | 100 | 606 |
|  | Musk1 | 2 | 166 | 476 |
|  | Musk2 | 2 | 166 | 6598 |
|  | Colon Cancer | 2 | 2000 | 62 |

These datasets were divided into three groups, including bi-class datasets, multiple-class datasets and high-dimensional datasets. In each bi-class dataset, the number of classes is two; in each multiple-class dataset, the number of classes ranges from 3 to 16; and in each high-dimensional dataset, the number of candidate features varies from 100 to 2000.

As a wrapper approach, the proposed algorithm adopted the SVM with a linear kernel [32] as a classifier. The pattern classification performance of a selected feature set was assessed by the percentage of correctly classified patterns obtained in the 10-fold cross-validation. It should be noted that other classifiers can also be used in the proposed approach and may lead to different performance. However, the discussion on optimal classifier selection is beyond the scope of this work.

The first group of experiments was performed on the nine bi-class datasets, in which the number of features ranges from 13 to 60 and the number of instances ranges from 208 to 4601. The parameters used in the proposed algorithm were listed in Table 2. Readers are referred to the Discussion Section for more details on parameter settings.

Table 2 Parameter settings in the proposed algorithm on the binary classification datasets

| Dataset | Number of Tribes $N_T$ | Tribe Size $N_{Tk}$ | Gaussian Mean $\mu_k$ | Standard Deviation $\sigma$ |
|---|---|---|---|---|
| WBCD | 3 | 600 | {2,5,8} | 0.75 |
| Heart | 3 | 600 | {3,7,11} | 1.08 |
| Australian | 3 | 600 | {3,7,11} | 1.17 |
| German | 3 | 600 | {5,11,17} | 1.75 |
| WDBC | 3 | 600 | {7,15,23} | 2.50 |
| Ionosphere | 3 | 600 | {8,17,26} | 2.83 |
| KR vs KP | 3 | 600 | {9,18,27} | 3.00 |
| Spam base | 3 | 600 | {14, 28, 32} | 4.75 |
| Sonar | 3 | 600 | {15, 30, 45} | 5.00 |

We compared the proposed TCbGA algorithm to several state-of-the-art feature selection algorithms, including DEMOFS [34], MOEA/D [35], MDisABC [36], W-QEISS [37], SB-ELM [38], HPSO-LS [39], MoDE [40], GASNCM [41], GCACO [42], GCNC [43], UFSACO [44], FFW-DGC [45], QIFS [46], FSFWISIW [47], BALO [48], MI-SC [49], VMBACO [50], HDBPSO [51] and bGWO [52]. Table 3 shows the mean and standard deviation of the classification accuracy obtained by applying our algorithm to each dataset 25 times and the performance of other algorithms reported in the literature. It reveals that TCbGA achieved the highest average classification accuracy on six datasets and the second highest average classification accuracy on the other two datasets. The results demonstrate that the proposed feature selection algorithm has substantially improved performance in bi-class pattern classification problems.

Table 3 Pattern classification accuracy (mean ± standard deviation) of different algorithms on eight bi-class datasets

| Dataset | Algorithm | Accuracy (%) | Dataset | Algorithm | Accuracy (%) |
|---|---|---|---|---|---|
| Spam Base | MoDE, 2015 | 91.83 ± 0.31 | Sonar | MoDE, 2015 | 84.18 ± 0.73 |
| | VMBACO, 2016 | 89.42 ± 1.44 | | FSFOA, 2016 | 82.69 |
| | GCACO, 2015 | 88.38 ± 1.33 | | GCACO, 2015 | 82.38 ± 1.51 |
| | GCNC, 2015 | 88.21 ± 1.15 | | GCNC, 2015 | 76.33 ± 2.52 |
| | UFSACO, 2014 | 87.92 ± 0.76 | | bGWO, 2016 | 73.1 |
| | MOEA/D, 2015 | 88.48 | | **HSA, 2016** | **85.2** |
| | **Proposed TCbGA** | **91.85 ± 0.09** | | Proposed TCbGA | 84.62 ± 0.03 |
| German | MOEA/D, 2015 | 71.3 | Australian | QIFS, 2017 | 85.52 ± 5.20 |
| | W-QEISS, 2016 | 76.0 | | MOEA/D, 2015 | 84.64 |
| | MDisABC, 2015 | 70.15 ± 1.87 | | **FFWDGC, 2017** | **88.09** |
| | **Proposed TCbGA** | **78.0 ± 0.05** | | W-QEISS, 2016 | 86.72 |
| | | | | Proposed TCbGA | 87.25 ± 0.06 |
| KR vs KP | BALO, 2016 | 96.7 | Heart | HPSO-LS, 2016 | 78.84 ± 2.07 |
| | bGWO, 2016 | 94.4 | | W-QEISS, 2016 | 84.3 |
| | FSFWISIW, 2015 | 94.12 | | bGWO, 2016 | 80.7 |
| | **Proposed TCbGA** | **99.40 ± 0.008** | | HSA, 2016 | 82.64 |
| | | | | **Proposed TCbGA** | **85.19 ± 0.06** |
| WDBC | HPSO-LS, 2016 | 98.27 ± 0.4 | Ionosphere | MoDE, 2015 | 93.68 ± 0.26 |
| | GASNCM, 2016 | 93.42 ± 2.0 | | UFSACO, 2014 | 88.61 ± 0.76 |
| | GCACO, 2015 | 94.14 ± 1.36 | | GCACO, 2015 | 90.41 ± 1.90 |
| | GCNC, 2015 | 95.34 ± 1.09 | | QIFS, 2017 | 86.69 ± 5.87 |
| | UFSACO, 2014 | 92.06 ± 0.77 | | MDisABC, 2015 | 93.62 ± 1.64 |
| | **Proposed TCbGA** | **98.78 ± 0.004** | | FSFOA, 2016 | 93.16 |
| | | | | **Proposed TCbGA** | **98.32 ± 0.04** |

To demonstrate the performance of TCbGA in multi-class classification problems, the second group of experiments was performed on the seven datasets, in which the number of features ranges from 13 to 279, the number of classes ranges from 3 to 16, and the number of instances ranges from 32 to 5000. The parameters used in TCbGA were listed in Table 4.

Table 4 Parameter settings in the proposed algorithm on seven multi-class datasets

| Dataset | Number of Tribes $N_T$ | Tribe Size $N_{Tk}$ | Gaussian Mean $\mu_k$ | Standard Deviation $\sigma$ |
|---|---|---|---|---|
| Wine | 3 | 600 | {3,7,11} | 1.08 |
| Zoo | 3 | 600 | {4,8,12} | 1.33 |
| Vehicle | 3 | 600 | {4,9,14} | 1.50 |
| Waveform | 3 | 600 | {5,11,17} | 1.75 |
| Dermatology | 3 | 600 | {8,17,25} | 2.75 |
| Lung | 3 | 600 | {14, 28, 42} | 4.67 |
| Arrhythmia | 6 | 2000 | {39,79,119, 159,199,239} | 13.29 |

Table 5 shows the mean and standard deviation of the classification accuracy obtained by applying our algorithm to each dataset 25 times and the performance of other algorithms reported in the literature. It reveals that the feature subset selected by TCbGA produces the highest average classification accuracy on six datasets and the second highest average classification accuracy on the other dataset. This experiment demonstrates that the proposed feature selection algorithm has substantially improved performance in multi-class pattern classification problems.

To assess the performance of TCbGA in selecting optimal features from a relative large set of features, the third group of experiments was carried out on four datasets, including Musk1, Musk2, Hill-Valley and Colon Cancer data, in which the number of features ranges from 100 to 2000 and the number of instances ranges from 62 to 6598. In this experiment, we divided the population into much more tribes than those in the first two groups of experiments. For instance, when applying our algorithm to the Colon Cancer dataset, which has 2000 features, we partitioned the population into 13 tribes, set the mean of the Gaussian distribution in these tribes to 136, 280, 424, 568, 712, 856, 1000, 1144, 1288, 1432, 1576, 1720 and 1864, respectively, and set the standard deviation to 47.62. Table 6 shows the parameters used on the four datasets in TCbGA.

Table 5 Pattern classification accuracy (mean ±standard deviation) of different algorithms on seven multi-class datasets

| Dataset | Algorithm | Accuracy (%) | Dataset | Algorithm | Accuracy (%) |
|---|---|---|---|---|---|
| Wine | GASNCM, 2016 | 95.37 ± 1.80 | Dermatology | VMBACO, 2016 | 95.17 ± 2.66 |
| | GCACO, 2015 | 94.09 ± 2.58 | | QIFS, 2017 | 95.32 ± 4.37 |
| | HPSO-LS, 2016 | 97.17 ± 1.39 | | FSFWISIW, 2015 | 95.28 |
| | GCNC, 2015 | 94.42 ± 1.04 | | FSFOA, 2016 | 96.99 |
| | VMBACO, 2016 | 99.10 ± 1.00 | | GCNC, 2015 | 88.21 ± 1.15 |
| | **Proposed TCbGA** | **99.60 ± 0.07** | | **Proposed TCbGA** | **99.65 ± 0.004** |
| Zoo | GASNCM, 2016 | 96.1 ± 3.50 | Arrhythmia | HPSO-LS, 2016 | 53 ± 2.28 |
| | MOEA/D, 2015 | 95.42 | | GCACO, 2015 | 60.51 ± 5.42 |
| | FFWDGC, 2017 | 98.02 | | GCNC, 2015 | 59.08 ± 2.38 |
| | bGWO, 2016 | 87.9 | | MOEA/D, 2015 | 65.77 |
| | **W-QEISS, 2016** | **99.22** | | UFSACO, 2014 | 59.22 ± 2.98 |
| | Proposed TCbGA | 98.03 ± 0.009 | | **Proposed TCbGA** | **74.80 ± 0.08** |
| Vehicle | GASNCM, 2016 | 73.37 ± 0.01 | Waveform | ABACO, 2015 | 79.8 ± 0.56 |
| | FFWDGC, 2017 | 75.6 | | MOEA/D, 2015 | 83.65 |
| | MDisABC, 2015 | 79.31 ± 1.85 | | BALO, 2016 | 80.0 |
| | HSA, 2016 | 80.07 | | bGWO, 2016 | 78.6 |
| | **Proposed TCbGA** | **86.05 ± 0.06** | | **Proposed TCbGA** | **85.43 ± 0.004** |

| Dataset | | | | | |
|---------|---|---|---|---|---|
| Lung | GASNCM, 2016 | 95.38 ± 3.97 | | | |
| | MDisABC, 2015 | 76.65 ± 4.36 | | | |
| | HSA, 2016 | 88.65 | | | |
| | **Proposed TCbGA** | **96.23 ± 0.003** | | | |

We compared the mean and standard deviation of the classification accuracy obtained by applying our algorithm to each dataset 25 times to the performance of several state-of-the-art solutions reported in the literature in Table 7. It shows that our algorithm achieves substantially higher accuracy than other approaches on these datasets. Moreover, the comparative results also suggest that TCbGA algorithm has a more obvious advantage over other feature selection methods when the number of candidate feature components is relatively large.

Table 6 Parameter settings in the proposed algorithm on four high-dimensional datasets

| Dataset | Number of Tribes $N_T$ | Tribe Size $N_{Tk}$ | Gaussian Mean $\mu_k$ | Standard Deviation $\sigma$ |
|---------|---|---|---|---|
| Hill-Valley | 3 | 1000 | {25, 50, 75} | 8.33 |
| Musk1 | 6 | 1000 | {24, 48, 72, 96,120,144} | 7.9 |
| Musk2 | 6 | 1000 | {24, 48, 72, 96,120,144} | 7.9 |
| Colon Cancer | 13 | 6000 | {136,280,424,568,712,856,1000, 1144,1288, 1432,1576,1720,1864} | 47.62 |

Table 7 Pattern classification accuracy (mean ± standard deviation) of different algorithms on four high-dimensional datasets

| Dataset | Algorithm | Accuracy (%) | Dataset | Algorithm | Accuracy (%) |
|---------|-----------|--------------|---------|-----------|--------------|
| Hill-Valley | DEMOFS, 2014 | 60.46 | Musk2 | FSFWISIW, 2015 | 93.3 |
| | MOEA/D, 2015 | 57.50 | | BALO, 2016 | 96.4 |
| | MDisABC, 2015 | 54.05 ± 2.05 | | MI-SC, 2016 | 91.9 |
| | **Proposed TCbGA** | **60.53 ± 0.01** | | **Proposed TCbGA** | **99.23 ± 0.05** |
| Colon Cancer | HDBPSO, 2015 | 90.28 ± 0.15 | Musk1 | MOEA/D, 2015 | 81.52 |
| | GCACO, 2015 | 81.42 ± 3.51 | | W-QEISS, 2016 | 76.02 |
| | GCNC, 2015 | 82.37 ± 1.79 | | MDisABC, 2015 | 85.29 ± 2.07 |
| | HPSO-LS, 2016 | 83.88 ± 4.09 | | BALO, 2016 | 89.2 |
| | **Proposed TCbGA** | **96.50 ± 0.02** | | **Proposed TCbGA** | **94.27 ± 0.08** |

Finally, the Friedman non-parametric test with a significant level of 0.05 [54] was employed to compare our proposed TCbGA algorithm against all the other methods that perform best on each dataset. In this statistic test, the null hypothesis $H_0$ affirms the equal behavior of the comparable methods. Hence, under $H_0$, each method possesses equal rank, which conforms that each method is equally efficient with others. The alternative hypothesis $H_1$ endorses the difference in performances among the comparable methods. The Friedman test we performed shows that the chi-square $(\mathcal{X}^2)$ value is 5.47 and the p-value is 0.0193. The p-value is smaller than the significance level 0.05. Meanwhile, the chi-square value is larger than critical value, which is 3.84 at 0.05 significance level and (2-1) = 1 degree of freedom. Hence, $H_0$ is rejected and $H_1$ is accepted. This result demonstrates that the performance improvement of the proposed algorithm is significant.

Table 8 gives the results of T-test. It shows that, when setting the significance level in the T-tests to 0.05, TCbGA performs significantly better than eight out of 15 start-of-the-art feature selection methods, including GCACO, GCNC, MOEA/N, GASNCM, UFSACO, MDisABC, HAS and bGWO on the datasets used in this study. Therefore, three groups of comparative experiments suggest that the proposed TCbGA algorithm is able to select better feature subset to improve significantly the accuracy of pattern classification on those 20 datasets.

Table 8 Results of T-test of 16 feature selection algorithms' accuracy on 20 datasets

| Algorithm | p-Value | Algorithm | p-Value |
|---|---|---|---|
| GCACO | 0.0080 * | FFW-DGC | 0.1028 |
| GCNC | 0.0172 * | QIFS | 0.1849 |
| MOEA/D | 0.0256 * | FSFOA | 0.1106 |
| GASNCM | 0.0319 * | MoDE | 0.1541 |
| UFSACO | 0.0029 * | W-QEISS | 0.0578 |
| MDisABC | 0.0220 * | VMBACO | 0.1636 |
| HSA | 0.0277 * | HPSO-LS | 0.2000 |
| bGWO | 0.0047* | TCbGA | / |

## 4. Discussion

### 4.1 Parameter Settings

As an evolutionary algorithm, the proposed feature selection method involves a number of parameters. Most of them, such as the size of population, maximum number of generations, crossover rate and mutation rate, are the commonly used parameters in traditional GA and can be set under the general guidelines for GA [55]. The number of tribes $N_T$ and the statistical parameters in each Gaussian distribution $\mathcal{N}(\mu_k, \sigma^2)$ are introduced to ensure that different groups of individuals can explore different parts of the solution space, and hence play an important role in this algorithm.

To ensure that the features that make up the global optimum do exist in a single tribe, every subspace $\Omega_m$ ($1 \leq m \leq N$) should be explored by one or two tribes. As $\{\mu_1, \mu_2, \cdots, \mu_{N_T}\}$ equally distributes in the range $[1, N]$, we let any two adjacent tribes be half-overlap and each tribe $T_k$ equally take care of $\frac{2}{N_T+1}$ of the solution space, shown as an example in Fig. 2. Hence, the part of solution space to be explored by the tribe $T_k$ can be denoted by $\bigcup_{m=m^-}^{m^+} \Omega_m$, where $m^- = \left\lceil \mu_k - \frac{N}{N_T+1} \right\rceil$ and $m^+ = \left\lfloor \mu_k + \frac{N}{N_T+1} \right\rfloor$. Since $T_k \cap \Omega_{m^-} \neq \emptyset$ and $T_k \cap \Omega_{m^+} \neq \emptyset$, the standard deviation $\sigma$, population size $N_P$ and number of tribes $N_T$ must satisfy the following constraint

$$\forall k, \ \mathrm{n}_{km^-} = \mathrm{n}_{km^+} \geq 1 \ . \tag{6}$$

Applying Eq. (2) to this constraint, we have

$$\frac{\frac{1}{\sigma\sqrt{2\pi}} exp\left[-\frac{(m^+ - \mu_K)^2}{2\sigma^2}\right]}{\sum_{i \in \aleph} \frac{1}{\sigma\sqrt{2\pi}} exp\left[-\frac{i^2}{2\sigma^2}\right]} \cdot N_{Tk} \geq \frac{1}{2} \ . \tag{7}$$

Another constraint on the standard deviation $\sigma$ is that the search scope of each tribe falls into the range $[\mu_k - 3\sigma, \mu_k + 3\sigma]$. Hence, we have

$$\sigma \geq \frac{N}{3(N_T+1)} \ . \tag{8}$$

Meanwhile, we notice that, when $\sigma > 0.7$,

$$\sum_{i \in \aleph} \frac{1}{\sigma\sqrt{2\pi}} exp\left[-\frac{i^2}{2\sigma^2}\right] \approx 1. \tag{9}$$

Thus, Eq. (7) can be rewritten as

$$\frac{1}{\sigma\sqrt{2\pi}}exp\left[-\frac{(m^+-\mu_\kappa)^2}{2\sigma^2}\right]\cdot N_{Tk} \geq \frac{1}{2} . \tag{10}$$

Here we define a function

$$\varphi(m) = \frac{1}{\sigma\sqrt{2\pi}}exp\left[-\frac{(m-\mu_\kappa)^2}{2\sigma^2}\right]\cdot N_{Tk} . \tag{11}$$

Then, Eq. (10) can be rewritten as

$$\varphi(m^+) \geq \frac{1}{2} . \tag{12}$$

Since $\varphi(m)$ decreases monotonously in the range $(\mu_k, +\infty)$ and $\mu_k < m^+ \leq \mu_k + 3\sigma$, we get

$$\varphi(m^+) \geq \varphi(\mu_k + 3*\sigma). \tag{13}$$

To ensure that $\varphi(m^+) \geq \frac{1}{2}$, we just need set

$$\varphi(\mu_k + 3*\sigma) \geq \frac{1}{2} . \tag{14}$$

Applying Eq. (11) to Eq. (14), we have

$$\frac{1}{\sigma\sqrt{2\pi}}exp\left[-\frac{((\mu_k+3*\sigma)-\mu_\kappa)^2}{2\sigma^2}\right]\cdot N_{Tk} \geq \frac{1}{2} . \tag{15}$$

Simplifying Eq. (15), we get

$$\sigma \leq \frac{2}{\sqrt{2\pi}} e^{-\frac{9}{2}} \cdot N_{Tk} \approx 0.008864 N_{Tk} . \tag{16}$$

Therefore, the standard deviation $\sigma$ is theoretically computed to take a value from the range $\left[\frac{N}{3(N_T+1)}, 0.008864 N_{Tk}\right]$.

Meanwhile, applying Eq. (8) to Eq. (16), we have

$$\frac{N}{3(N_T+1)} \leq \frac{2}{\sqrt{2\pi}} e^{-\frac{9}{2}} \cdot N_{Tk}. \tag{17}$$

Hence

$$N_T \geq \frac{\sqrt{2\pi}N}{6e^{-\frac{9}{2}}N_{Tk}} - 1 \approx 37.6066\frac{N}{N_{Tk}} - 1 . \tag{18}$$

Since the inter-tribe competition may not work when $N_T \leq 2$, we suggest to set the number of tribes $N_T$ to $\max\left(3, \left\lceil 37.6066\frac{N}{N_{Tk}} - 1\right\rceil\right)$.

We also chose the WDBC, Lung, Dermatology and Musk1 datasets as a case study to investigate the setting of the number of tribes $N_T$. The number of features in those datasets is 30, 56, 33 and 166, respectively. The size of tribe is set to 600 for WDBC, Lung and Dermatology, and 1000 for Musk1 due to a relatively large solution space. According to the above theoretical analysis, the optimal number of tribes should be 3, 3, 3 and 6, respectively. Fig. 5 plots the variation of the classification accuracy over the number of tribes. It shows that the estimated optimal number of tribes, marked by dotted lines in this figure, is highly consistent with our computational results.

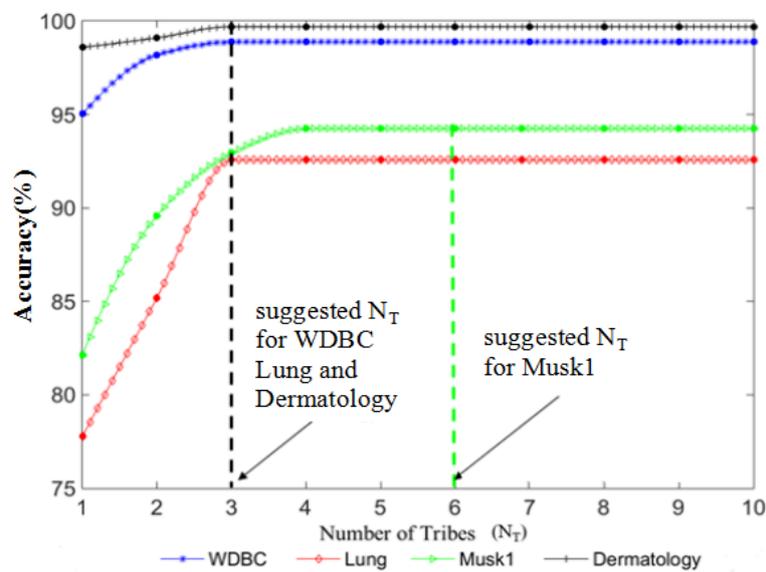

Fig.5. Variation of classification accuracy over the number of tribes on four datasets

**4.2 Tribe Competition**

A distinct feature of the proposed algorithm is to incorporate tribe competition into the evolution process. Since we divided the solution space into several half-overlapped subspaces, each being explored by a tribe, the features that make up the global optimum must exist in the subspaces searched by one or two tribes, which are named as the elite tribes. Ideally, we should keep only elite tribes and allocate them all individuals. Unfortunately, we do not know which tribe is the elite. Hence, we performed the inter-tribe competition and regarded the winners as the elite tribes. Then, we enlarged the size of the predicted elite tribes to enable them to have more search power and cut down the size of those defeated tribes to save computing resource. This penalty and award strategy enables the algorithm not only to search the solution space locally, but also to quickly look for the global optimal. The characteristic of human tribes in the primitive society, i.e. tribes evaluate themselves to get more adaptive ability and plunder other weaker tribes of their resources to make themselves stronger, contributes to better understand this competition strategy.

Fig.6 illustrates the change of the size of three tribes during the evolution and competition when applying the proposed algorithm to the Dermatology dataset. It shows that, although the population size is maintained, the sizes of tribes change dynamically. Our result indicates that the highest classification accuracy is achieved when using a subset of 23 features. This optimal solution lies exactly in the subspace that is explored by the third tribe, in which the average number of selected features is 25. This result is completely consistent with our observation of the increase of third tribe's size.

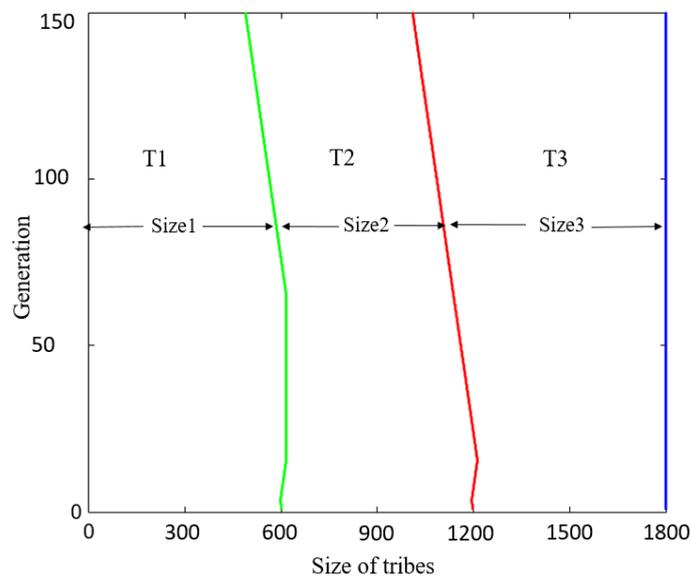

Fig .6. Change of the size of three tribes (denoted by $T_1$, $T_2$ and $T_3$) during the evolution

However, our prediction of the elite tribe is not always correct, since it may perform worse than others at the early stage of evolution, due to the complexity of feature selection problems.

Fig.6 also reveals that the size of each tribe is not changed monotonically and the elite tribe $T_3$ was even considered to be the worst one in early competitions. Therefore, we chose the most conservative implementation of the penalty and award strategy that is to enlarge and shrink only the best and worst tribe by one, although there are different implementations, including enlarging and/or shrinking one or more tribes by adding or removing one or more individuals each time. More aggressive implementations may speed up the convergence of the evolution, but at the risk of missing the elite tribe due to shrinking it too much at a too early stage.

Another important issue related to the inter-tribe competition is how frequent it should be performed. Generally, more frequent competition gives the algorithm more opportunity to adjust the size of tribes, whereas less frequent competition gives each tribe more opportunity to find a better solution before its size is adjusted.

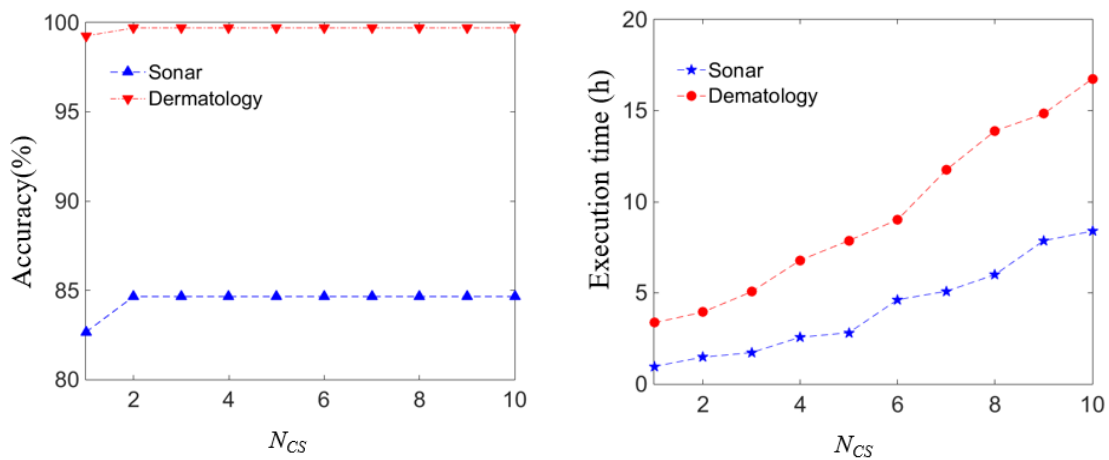

Fig.7 Accuracy (left) and time-cost (right) of our algorithm on two datasets over different settings

We adopted the Sonar and Dermatology datasets as a case study to explore the impact of the frequency of inter-tribe competitions on the algorithms' performance. The results were shown in Fig. 7, in which $N_{CS}$ stands for performing the inter-tribe competition after every $N_{CS}$ generations. It reveals that setting $N_{CS}$ to 1 decreases the classification accuracy on both datasets; whereas using $N_{CS}$ greater than 2 results in almost the same accuracy but significantly increased time-cost. The reason for the increased time-cost lies in the fact that a larger $N_{CS}$ makes the algorithm converge more slowly and require more generations of evolution to achieve a satisfying result. Therefore, considering both the accuracy and complexity, we set $N_{\text{CS}} = 2$.

**4.3 Robustness**

The proposed method employs heuristic-guided stochastic search, and hence may produce different near optimal solutions in multiple runs. Table 9 gives average performance obtained by applying this algorithm with random initializations to those 20 datasets 25 times. In this table, $AC_{\text{ave}}$ and $AC_{\text{std}}$ are the mean and standard deviation of the classification accuracy, respectively, $AS_{\text{num}}$ is the average number of selected features, and $Total_{\text{num}}$ is the total number candidate feature component to be selected. The very small standard deviation of accuracy shown in this table demonstrates that the proposed algorithm is relatively robust to initializations.

Table 9 Average performance of the proposed algorithm on 20 datasets for 25 runs

| Dataset | $AC_{ave}$ | $AC_{std}$ | $AS_{num}/Total_{num}$ |
|---|---|---|---|
| *WBCD* | 98.09% | 2.13e-03 | 5.2/9 |
| Heart | 85.19% | 5.81 e-02 | 4.2/13 |
| Australian | 87.25% | 6.23 e-02 | 6.4/14 |
| German | 78% | 5.32 e-02 | 12.3/21 |
| *WDBC* | 98.78% | 3.85e-03 | 20.8/30 |
| Ionosphere | 98.32% | 3.52e-02 | 14.2/34 |
| *KR vs KP* | 99.40% | 7.85e-03 | 26.0/36 |
| Lung | 96.23% | 2.79e-03 | 9.2/56 |
| Spam base | 91.85% | 8.93e-02 | 19.0/57 |
| Sonar | 84.62% | 2.75e-02 | 8.7/60 |
| Hill-Valley | 59.23% | 1.27e-02 | 38.2/100 |
| Musk1 | 94.27% | 7.52e-02 | 97.3/166 |
| Musk2 | 99.23% | 4.52e-02 | 85.5/166 |
| Colon cancer | 96.50% | 2.32e-02 | 19.3/2000 |
| Wine | 99.60% | 7.23e-03 | 9.0/13 |
| Zoo | 98.03% | 8.67 e-03 | 5.1/16 |
| Waveform | 85.43% | 3.85e-03 | 18.0/21 |
| Vehicle | 86.05% | 5.73e-02 | 12.5/18 |
| Dermatology | 99.65% | 4.25e-03 | 24.0/33 |
| Arrhythmia | 74.80% | 7.56e-02 | 34.7/279 |

**4.4 Computational Complexity**

As a wrapper feature selection method and an evolutionary approach, the proposed algorithm has a relatively high computational complexity. Most computation is spent on the evaluation of individuals' fitness. Given the size of population $N_P$ and maximum number of generation $N_G$, such evaluation is performed $N_P \cdot N_G$ times. Each time, the feature subset specified by an individual is used to train a classifier and test its accuracy via cross validation. Therefore, the computational complexity is also determined by the number of candidate features, number of classes and number of instances. More instances and classes usually require more computation, and more candidate features may lead to large selected feature subset and thus also require increased computation. Table 10 give the average time cost of training the model and testing one instance on seven datasets (Intel Core i7-4790 CPU 3.2GHz, NVidia GTX Titan X GPU, 32GB memory and Matlab Version 2014). Although the off-line training is extremely time-consuming, the proposed algorithm has the ability to select the optimal features for various classification problems. We believe the ever increase of computational power, particularly the prevalence of GPU-based parallel computation will make it more computationally attractive. Meanwhile, during online testing, applying the selected optimal feature subset to solving pattern classification problems is very efficient.

Table 10 Average time-cost of the proposed algorithm on 20 datasets

| Dataset | Train (h) | Test (s) | Dataset | Train (h) | Test (s) |
| --- | --- | --- | --- | --- | --- |
| WBCD | 0.21 | 0.01 | Hill-Valley | 4.88 | 0.59 |
| Heart | 3.18 | 0.12 | Musk1 | 5.62 | 0.68 |
| Australian | 5.68 | 0.52 | Musk2 | 8.53 | 1.23 |
| German | 4.73 | 0.35 | Colon cancer | 5.48 | 0.52 |
| WDBC | 2.23 | 0.24 | Wine | 2.05 | 0.17 |
| Ionosphere | 2.97 | 0.35 | Zoo | 1.57 | 0.20 |
| KR vs KP | 9.25 | 1.05 | Waveform | 10.21 | 1.30 |
| Lung | 1.21 | 0.59 | Vehicle | 4.72 | 0.13 |
| Spam Base | 11.23 | 0.80 | Dermatology | 4.20 | 0.32 |
| Sonar | 1.52 | 0.29 | Arrhythmia | 7.22 | 0.86 |

**4.5 Feasibility**

The proposed algorithm has the ability to search the global optima, due to using the stochastic evolutionary strategy. However, it cannot guarantee to converge to the global optimal within limited generations. As a result, when the number of candidate features is small, it may have little advantage over other approaches. We used the WBCD dataset as a case study, where there are 699 instances from two classes and each instance consists of nine features. Since there are only $2^9 - 1$ possible feature subsets in this problem, we can find the optimal feature subset with respect to different classifiers via exhaustive searching. Table 11 give the classification accuracy and time cost of the proposed algorithm and exhaustive searching. It shows that the proposed algorithm can achieve the optimal classification accuracy, but may spend even more time than exhaustive searching. Therefore, when the number of candidate feature is too small, it may not be necessary to utilize multiple tribes to search different parts of the solution space, particularly when the solution space can be explored exhaustively.

Table 11 Classification accuracy and time cost of different approaches on the WBCD dataset

| Classifier | Feature Selection | Accuracy (%) | Time Cost (s) |
|---|---|---|---|
| SVM | Exhaustive Search | **98.09** | 19.25 |
| | Proposed | **98.09** | 761.08 |

On the contrary, if the number of candidate features is huge, we have to either use a large number of tribes ($N_T \geq \left\lceil 37.6066 \frac{N}{N_{Tk}} - 1 \right\rceil$) or set the standard deviation $\sigma$ to a large value. The former increases the computationally complexity; whereas the later makes the algorithm not to be able to focus on a small solution space and hence may produce less accuracy results, unless we set the tribe size to an even larger number that satisfies $\sigma < 0.008864 N_{Tk}$. Meanwhile, a huge number of instances also makes the proposed algorithm computationally intractable. In this case, we may use a randomly sampled small training set to evaluate the fitness of each individual. The discrepancy caused by this is determined by the generalization ability of the classifier used in the wrapper method.

## 5. Conclusion

In this paper, we propose the TCbGA algorithm for feature selection in pattern classification, which divides the population into multiple tribes, each containing a cohort of individuals. We encode each individual as a binary string to represent a possible feature selection scheme and modify the initialization and evolutionary operations to ensure that the number of selected features in each tribe follows a Gaussian distribution. Besides evolving each tribe independently, we introduce tribe competition to allow the tribe elite individuals to have increased searching power. Our results suggest that the proposed algorithm outperforms several state-of-the-art feature selection approaches on 20 benchmark datasets. Our future work will focus on incorporating classifier selection into the optimization process, reducing the complexity and extending this algorithm to solve feature selection problems with a super large feature set.

**Acknowledgement**

This work was supported in part by the National Natural Science Foundation of China under Grants 61471297 and 61231016, and in part by the Returned Overseas Scholar Project of Shaanxi Province, China. We appreciate the assistance we received by using the UCI Machine Learning Repository (http://mlr.cs.umass.edu/ml/).